\documentclass[letterpaper]{article} 
\usepackage[preprint]{aaai2027}  
\usepackage[hyphens]{url}  
\usepackage{graphicx} 
\usepackage{natbib} 
\usepackage{caption} 
\usepackage{booktabs}
\usepackage{algorithm}
\usepackage{algorithmic}
\graphicspath{{figures/}}

\newcommand{\metricmean}[1]{%
  \makebox[2.25em][r]{\ensuremath{#1}}}
\newcommand{\statunc}[1]{%
  \raisebox{-0.30ex}{%
    {\fontsize{7pt}{7pt}\selectfont\ensuremath{\pm #1}}}}
\newcommand{\meanstd}[2]{%
  \mbox{\metricmean{#1}\hspace{0.04em}\statunc{#2}}}
\newcommand{\bestmeanstd}[2]{%
  \mbox{\metricmean{\mathbf{#1}}\hspace{0.04em}\statunc{#2}}}

\title{ReOrder-OPD: Reliability-Aware Prompt Ordering for On-Policy Distillation}
\author{
    Ximo Zhu\textsuperscript{\rm 1}\equalcontrib,
    Ruiqi Liu\textsuperscript{\rm 2,3}\equalcontrib,
    Rong Wang\textsuperscript{\rm 2},
    Ping Wu\textsuperscript{\rm 2},
    Xiang Zheng\textsuperscript{\rm 2},
    Wenzhuo Xu\textsuperscript{\rm 2},
    Xubin Yao\textsuperscript{\rm 2},
    \\
    Zhiyuan Yan\textsuperscript{\rm 4},
    Bo Li\textsuperscript{\rm 1},
    Jun Gao\textsuperscript{\rm 1},
    Xiaolei Lv\textsuperscript{\rm 1}\corresponding
}
\affiliations{
    \textsuperscript{\rm 1}Hello Group Inc.\\
    \textsuperscript{\rm 2}Institute of Automation, CAS\\
    \textsuperscript{\rm 3}School of Advanced Interdisciplinary Sciences, UCAS\\
    \textsuperscript{\rm 4}Peking University
}

\begin{document}

\maketitle

\begin{abstract}
On-policy distillation (OPD) applies token-level teacher supervision to
student-generated trajectories, but this supervision is not always reliable.
Existing methods use local confidence or teacher--student agreement to weight,
filter, or truncate the sampled trajectory. These signals do not directly
determine whether the teacher can continue a student prefix to a correct
answer, and trajectory-level interventions can conflate one rollout's
unreliability with low expected training value of its prompt. We define
prompt-level teacher continuation reliability $R$ as the teacher's probability
of reaching a correct answer from a student prefix, averaged over prefixes and
trajectories induced by the current student. Oracle experiments show that
high-$R$ prompts yield larger OPD gains and that descending-$R$ training
outperforms random and ascending orders on a fixed prompt pool. Because
estimating $R$ requires many teacher continuations, we use the maximum ROUGE-5
F1 between one independent student rollout and verifier-correct same-prompt
teacher trajectories. Across ten equal-frequency bins of this actual score,
mean $R$ rises monotonically, showing that the proxy separates coarse
reliability levels. ReOrder-OPD sorts prompts by the proxy, then draws
independent on-policy training trajectories for vanilla OPD. It improves every
matched aggregate comparison across Qwen3 and Gemma4 mathematics settings and
Qwen3 code settings. Gains in all six FiRe-OPD and ExOPD settings show
that prompt ordering complements within-trajectory supervision.
\end{abstract}

\section{Introduction}

On-policy distillation (OPD) queries the teacher distribution along trajectories
generated by the current student. By supervising prefixes that the student is
likely to encounter at inference, OPD reduces the train--test prefix mismatch
of offline distillation. The teacher's continuation from a student-generated
prefix can nevertheless end with an incorrect answer, making the resulting
supervision unreliable.

\begin{figure}[t]
\centering
\includegraphics[width=\columnwidth]{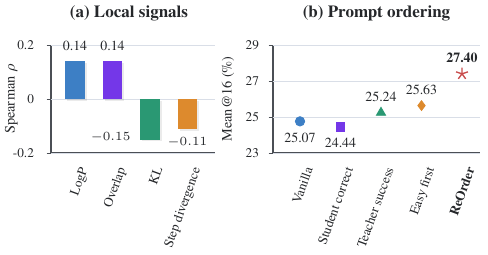}
\caption{Motivation studies. (a) Prompt-level Spearman correlations of local
signals with teacher continuation success on 905 prompts. (b) DeepMath-1K
AIME25/26 mean@16 under alternative prompt orders with the prompt pool,
objective, and budget fixed. Markers and labels show means over three seeds.}
\label{fig:motivation}
\end{figure}

Existing methods address this unreliability with teacher confidence,
token-level KL, top-$k$ overlap, entropy, or step-level divergence. These
quantities describe local confidence or teacher--student consistency along the
sampled trajectory; they do not directly answer whether the teacher can
continue the current prefix to a correct outcome. Figure~\ref{fig:motivation}(a)
shows that they are weakly associated with continuation success. Filtering,
truncation, and weighting then adjust supervision only within that observed
trajectory. Such interventions can inadvertently treat an unreliable rollout
as evidence that its prompt has low training value, thereby underestimating the
prompt's expected value for OPD. The student evolves throughout OPD, so the
same prompt can induce different trajectories and updates depending on when it
is visited. With the prompt pool, OPD objective, and update budget fixed,
changing only the prompt order produces clear performance differences, as
shown in Figure~\ref{fig:motivation}(b).

\begin{figure*}[t]
\centering
\includegraphics[width=\linewidth]{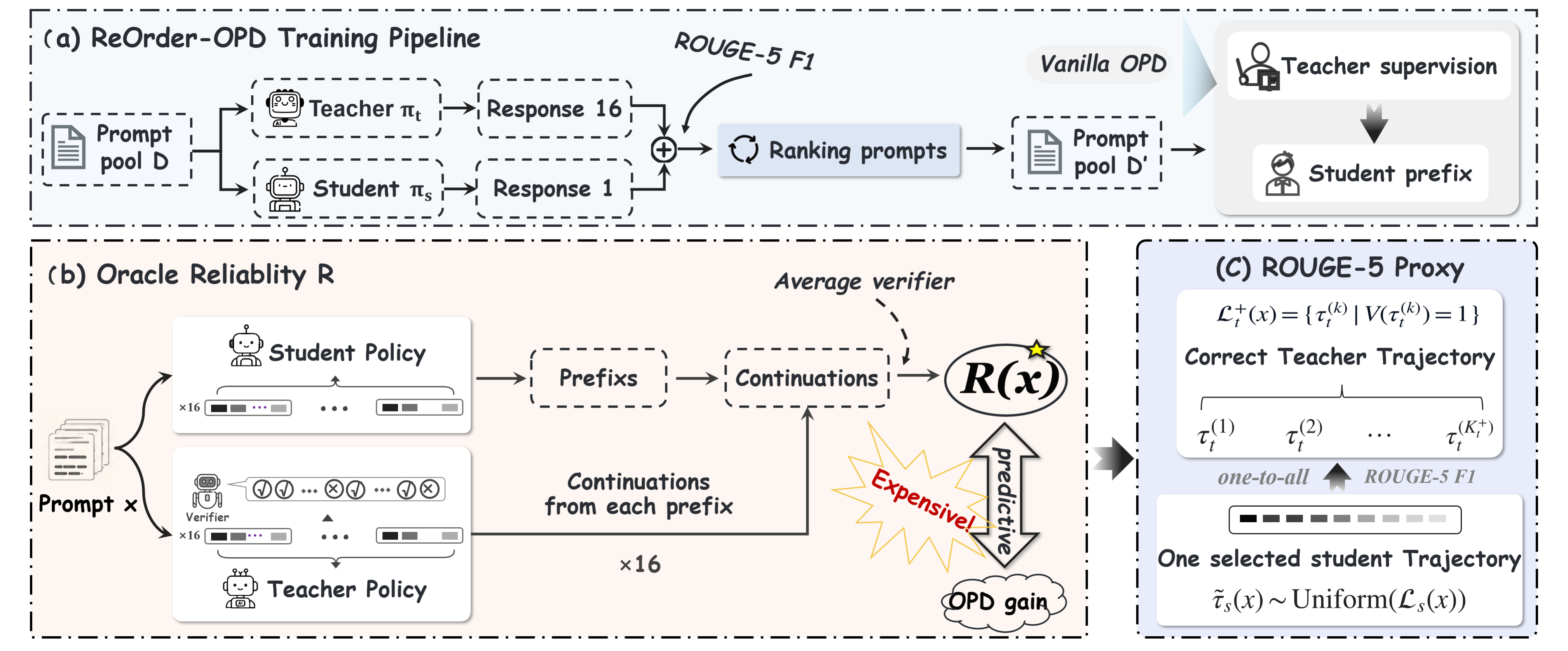}
\caption{Overview of ReOrder-OPD. (a) One student rollout and a
verifier-correct teacher library determine prompt order; training uses fresh
on-policy rollouts. (b) Oracle $R(x)$ averages teacher continuation outcomes
over prefixes and student trajectories and is used only for analysis. (c) The
practical score is maximum full-response ROUGE-5 F1 against correct
same-prompt teacher trajectories.}
\label{fig:overview}
\end{figure*}

We characterize outcome-level reliability by the probability that the teacher
continues a student prefix to a correct answer, then aggregate this probability
over prefixes and trajectories induced by the current student policy. The
resulting prompt-level teacher continuation reliability $R$ connects the two
decisions above. In oracle experiments, high-$R$ prompts produce larger OPD
gains than low-$R$ prompts, and descending-$R$ training outperforms random and
ascending orders while retaining the complete prompt pool. Because $R$
averages over the student's trajectory distribution, it provides the
prompt-level quantity required for scheduling.

Estimating $R$ requires repeated teacher continuations from many student
prefixes, whereas ordering needs only a coarse ranking. We instead score one
independent student trajectory by its maximum full-response ROUGE-5 F1 against
verifier-correct same-prompt teacher trajectories \citep{lin2004rouge}. Across
ten equal-frequency bins of this score, mean $R$ rises monotonically from 0.29
to 0.98, providing the coarse reliability levels needed for ordering.

ReOrder-OPD sorts the pool by this proxy, then draws a fresh on-policy rollout
when each prompt is visited. It changes prompt access order without
reusing the scoring rollout. FiRe-OPD and ExOPD instead determine how the
generated rollout is supervised, allowing the two intervention levels to be
combined directly.

Our contributions are:
\begin{itemize}
    \item We define outcome-level teacher continuation reliability and
    aggregate it into prompt-level $R$. Oracle subset and full-pool ordering
    experiments connect $R$ to OPD gains and motivate its use for prompt
    scheduling.
    \item We propose a maximum full-response ROUGE-5 proxy based on one
    independent student rollout. A monotonic trend across ten equal-frequency
    groups shows that it distinguishes coarse reliability strata.
    \item We introduce ReOrder-OPD as a prompt-level scheduling layer and
    demonstrate consistent matched aggregate gains across model families,
    student scales, random seeds, mathematics, and code. Data-scale
    experiments extend the evaluation to larger prompt pools, and gains on
    top of FiRe-OPD and ExOPD show that prompt ordering complements
    within-trajectory supervision across both objectives and at every tested
    scale.
\end{itemize}

\section{Related Work}

\subsection{On-Policy Distillation}

Classical knowledge distillation transfers knowledge by training a student to
match a teacher's output distribution
\citep{hinton2015distilling,gou2021knowledge}.
Sequence-level distillation further uses complete teacher-generated sequences
as training targets \citep{kim2016sequence}. For autoregressive models, the
student visits prefixes formed by its own outputs at inference time, so
training only on fixed teacher trajectories creates a distribution shift
between training and inference
\citep{ross2011reduction,bengio2015scheduled}. GKD and MiniLLM compute teacher
supervision on student-generated trajectories, establishing the basic form of
on-policy distillation for large language models
\citep{agarwal2024policy,gu2024minillm}.

Recent work further examines when this conditional supervision is effective.
Rethinking OPD relates training outcomes to teacher--student reasoning-pattern
agreement, new capabilities supplied by the teacher, and alignment on
high-probability tokens; Revisiting OPD identifies supervision imbalance in
sampled-token OPD and failures of teacher guidance on student-generated
prefixes \citep{li2026rethinking,fu2026revisiting}. These studies show
that teacher capability alone does not fully determine OPD performance;
supervision quality also depends on the reasoning states visited by the
student. Our measure asks whether the teacher can continue from those states
to a correct answer and tests how this outcome relates to prompt-level OPD
training value.

\subsection{Reliability-Aware OPD}

Existing methods primarily identify or handle unreliable supervision within
the current rollout. Prune-OPD uses top-$k$ compatibility to reweight and
truncate trajectories after drift, FiRe-OPD filters trajectories and then
softly reweights tokens, and PW-OPSD diagnoses branch viability to assign
position-dependent teacher-token weights; SOD adapts distillation strength
using step-level divergence
\citep{yang2026prune,li2026filter,liu2026teacher,zhong2026sod}. At the trajectory and
generation levels, Guided-OPD introduces teacher intervention during
multi-turn interaction, PG-OPD allocates continuation budgets according to
teacher--student overlap, and RG-OPD combines verifier feedback with
likelihood differences to gate distillation
\citep{li2026policy,zhao2026prefix,akhondzadeh2026reward}. G-OPD and its
ExOPD instance adjust the distillation signal from the objective side
\citep{yang2026learning}.

These methods decide how to supervise a trajectory that has already been
sampled. ReOrder-OPD instead uses an outcome-level prompt statistic to make
the earlier decision of when a prompt enters training. The selected OPD
objective then supervises an independently sampled on-policy trajectory,
which permits direct composition with methods such as FiRe-OPD and ExOPD.

\subsection{Curriculum Learning and Sample Ordering}

Curriculum learning changes optimization by arranging training samples.
Classical methods commonly organize data according to predefined difficulty
or the learner's current competence
\citep{bengio2009curriculum,kumar2010self,
platanios2019competence,hacohen2019power,soviany2022curriculum}.
Learned curricula instead use progress signals or a teacher or mentor policy
\citep{graves2017automated,jiang2018mentornet}.
In reasoning training, CDAS
selects samples according to the match between student capability and problem
difficulty \citep{kong2026rethinking}; further comparisons find that neither
easy-to-hard nor hard-to-easy order is universally superior
\citep{jia2026makes}.
Concurrent SEAD gates easy-to-hard prompt exposure using student competence
\citep{lee2026sead}. ReOrder instead retains the fixed prompt pool and orders
it by a teacher-continuation proxy computed from responses generated by the
current student.

ReOrder-OPD is an outcome-conditioned prompt curriculum for OPD. Rather than
ranking prompts by intrinsic complexity or student correctness, it ranks them
by how reliably the teacher can continue from states produced by the current
student. Static ReOrder-OPD computes this score once with the initial student
and delays rather than removes low-scoring prompts within the fixed pool.

\section{Preliminaries}

\paragraph{Problem setup, notation, and verifier.}
Let $\mathcal{D}$ be a set of reasoning prompts, let
$x\in\mathcal{D}$ denote one prompt, let $\pi_t$ be a fixed teacher, and let
$\pi_s^{(k)}$ be the student policy used to collect rollouts at update $k$.
Given $x$, the student samples a trajectory
$y^s=(y^s_1,\ldots,y^s_T)\sim\pi_s^{(k)}(\cdot\mid x)$. At position $t$,
$h_t=(x,y^s_{<t})$ denotes the prompt and the content previously generated by
the student. OPD queries teacher and student next-token distributions on these
student-generated prefixes. A deterministic answer verifier
$V(x,y)\in\{0,1\}$ evaluates the final answer of a complete response $y$, but
ignores intermediate reasoning and requires no reference derivation.

\paragraph{Vanilla on-policy distillation.}
We use sampled-token OPD. For a student token $y^s_t$ sampled at $h_t$, the
detached token-level distillation advantage $A_t$ is the difference between
teacher and sampling-student log-probabilities, and the policy ratio is
\begin{equation}
\begin{array}{c}
A_t=\log\pi_t(y^s_t\mid h_t)-\log\pi_s^{(k)}(y^s_t\mid h_t),\\[3pt]
r_t(\theta)=
\displaystyle\frac{\pi_\theta(y^s_t\mid h_t)}
{\pi_s^{(k)}(y^s_t\mid h_t)},\\[3pt]
\bar r_t(\theta)=
\mathrm{clip}\!\left(r_t(\theta),1-\epsilon,1+\epsilon\right).
\end{array}
\end{equation}
Here, $\pi_\theta$ is optimized, $\epsilon>0$ is the clipping threshold, and
the expectation is over sampled prompts and trajectories collected with
$\pi_s^{(k)}$. The clipped surrogate loss is
\begin{equation}
\mathcal{L}_{\mathrm{OPD}}(\theta)=
-\mathrm{E}\!\left[
\frac{1}{T}\sum_{t=1}^{T}
\min\!\left(r_t(\theta)A_t,\bar r_t(\theta)A_t\right)
\right].
\end{equation}

\section{Method}

\paragraph{Overview.}
Oracle estimates of prompt-level $R$ are used only for analysis and proxy
validation. ReOrder-OPD orders prompts with one student rollout, then draws
fresh on-policy rollouts for OPD updates; its dynamic extension refreshes only
the unvisited queue.

\subsection{Teacher Continuation Reliability}

Given a prompt $x$, the student policy $\pi_s$ generates a complete response
$y^s$. Let $z$ be a nonterminal prefix selected from $y^s$, excluding the
prompt. The teacher generates a continuation $c$ conditioned on $x$ and $z$,
producing $z\oplus c$. Its probability of reaching a correct answer is
\begin{equation}
R_{\mathrm{pref}}(x,z)=
\mathrm{E}_{c\sim\pi_t(\cdot\mid x,z)}
\left[V(x,z\oplus c)\right].
\end{equation}
Let $G(y^s)$ denote the rule for sampling nonterminal prefixes. Prompt-level
teacher continuation reliability is
\begin{equation}
R(x;\pi_s,\pi_t)=
\mathrm{E}_{y^s\sim\pi_s(\cdot\mid x)}
\left[
\mathrm{E}_{z\sim G(y^s)}
\left[R_{\mathrm{pref}}(x,z)\right]
\right].
\label{eq:reliability-population}
\end{equation}

In practice, we sample $K_s$ student trajectories, select $N$ prefixes from
each trajectory, and draw $M$ teacher continuations from each prefix. If
$S(x)$ of the resulting $K_sNM$ complete responses are verifier-accepted,
\begin{equation}
\widehat{R}(x)=\frac{S(x)}{K_sNM}.
\label{eq:reliability}
\end{equation}
$R$ is jointly determined by the prompt, student policy, and teacher.
$R_{\mathrm{pref}}$ measures outcome-level reliability at one prefix, while
the outer expectations in Equation~\ref{eq:reliability-population} aggregate
over prefixes and trajectories induced by the current student. We estimate
$R$ only for oracle diagnosis and proxy validation. This joint dependence is
intentional: scheduling needs the expected reliability of teacher supervision
under states visited by the current student, rather than an intrinsic prompt
property isolated from the student or teacher. Practical ReOrder-OPD uses the
proxy defined next.

\subsection{From Reliability to a One-Rollout Proxy}

Prompt ordering does not require a calibrated estimate of $R$ for every
prompt; it requires a score that separates broad reliability levels. We use
a trajectory-level proxy grounded in verified outcomes.
Local confidence and distributional-agreement signals characterize next-token
behavior along an observed trajectory, whereas our proxy compares one complete
student response with verified-correct solutions to the same prompt. It
requires no teacher continuations from intermediate student prefixes during
scoring.

For every prompt, we sample $K_t$ teacher rollouts under a fixed configuration
and retain only responses with correct final answers:
\begin{equation}
\mathcal{C}_t(x)=
\left\{
y_i^t:
y_i^t\sim\pi_t(\cdot\mid x),\
V(x,y_i^t)=1,\
1\leq i\leq K_t
\right\}.
\end{equation}
For a student policy $\pi_s$, we independently sample exactly one complete
trajectory $\widetilde{y}^s(x;\pi_s)$ before similarity is computed. Its
selection does not depend on ROUGE-5. The proxy is
\begin{equation}
q(x;\pi_s)=
\max_{y^t\in\mathcal{C}_t(x)}
\mathrm{ROUGE\mbox{-}5\,F1}\!\left(\widetilde{y}^s(x;\pi_s),y^t\right).
\label{eq:proxy}
\end{equation}

The maximum accommodates multiple valid solution styles by retrieving the
closest verified successful trajectory for the same prompt. ROUGE-5 is
computed on complete responses, including the final answer and excluding the
prompt. Exactly one student rollout is scored; if
$\mathcal{C}_t(x)=\emptyset$, the prompt is placed after all scored prompts.
The resulting $q$ supplies an ordinal priority for ordering.

\subsection{ReOrder-OPD}

Static ReOrder-OPD evaluates Equation~\ref{eq:proxy} once with the initial
student $\pi_s^{(0)}$, sorts scored prompts from high to low, and appends
unscored prompts. No prompt is removed by its score; the configured update
budget determines how much of the fixed queue is consumed. ReOrder-OPD then
performs vanilla OPD on this queue. To preserve the order, ReOrder-OPD disables
data shuffling; vanilla OPD shuffles randomly.

When training reaches a prompt, the current student generates a fresh
on-policy rollout; the scoring rollout determines only its queue position.
Apart from order, matched runs retain the same teacher, initialization, prompt
pool, rollout procedure, loss, optimizer, batch size, and update count.

This separation locates ReOrder at the prompt-scheduling level. Vanilla OPD
is the main downstream objective; alternatively, a trajectory-level objective
such as FiRe-OPD or ExOPD can supervise the same training rollout without
changing the queue construction.

\paragraph{Why order matters in an on-policy objective.}
High-$R$ prompts are more likely to provide teacher supervision whose
continuation reaches a correct answer from student-visited states. Placing
them earlier gives the initial OPD updates more reliable teacher targets.
At update $k$, both the sampled trajectory and the teacher targets are
conditioned on the current student:
\begin{equation}
\begin{array}{c}
y_k^s\sim\pi_s^{(k)}(\cdot\mid x_k),\\[2pt]
\pi_s^{(k+1)}=\mathcal{U}(\pi_s^{(k)};x_k,y_k^s,\pi_t),
\end{array}
\label{eq:update-path}
\end{equation}
where $\mathcal{U}$ denotes one OPD update. Presenting prompt
$a$ before $b$ changes the policy that generates the trajectory for $b$; in
general, $\mathcal{U}_b(\mathcal{U}_a(\pi))$ and
$\mathcal{U}_a(\mathcal{U}_b(\pi))$ need not coincide. ReOrder-OPD uses this
path dependence to visit higher coarse reliability strata first. Early
reliable updates alter the policy that generates later trajectories and
teacher targets, propagating the effect of prompt order through training.

\subsection{Dynamic Extension}

Static ReOrder-OPD is the main method. Because $q(x;\pi_s)$ depends on the
current student, its initial ranking can become less informative as training
progresses. Dynamic ReOrder-OPD uses a set of refresh updates
$\mathcal{T}$. At each $t\in\mathcal{T}$, it draws one new trajectory from the
current student for every prompt still in the queue, recomputes
Equation~\ref{eq:proxy}, and re-sorts only those unvisited prompts. The correct
teacher libraries remain fixed, and prompts already used for an update never
return to the queue. Algorithm~\ref{alg:reorder} gives both variants; static
ReOrder-OPD is the special case $\mathcal{T}=\emptyset$.

\begin{algorithm}[ht]
\caption{Static and dynamic ReOrder-OPD}
\label{alg:reorder}
\begin{algorithmic}[1]
\REQUIRE Prompt set $\mathcal{D}$; correct teacher libraries
$\{\mathcal{C}_t(x)\}$; initial student $\pi_s^{(0)}$; batch size $B$;
update budget $U$; refresh updates $\mathcal{T}$
\STATE $\mathcal{Q}\leftarrow\mathcal{D}$; $k\leftarrow0$
\STATE Draw one rollout for every $x\in\mathcal{Q}$; score prompts with a
nonempty library by Equation~\ref{eq:proxy}, sort them descending, and append
unscored prompts
\WHILE{$k<U$ and $\mathcal{Q}\neq\emptyset$}
    \IF{$k\in\mathcal{T}$ and $k>0$}
        \STATE Draw one current-student rollout for every $x\in\mathcal{Q}$;
        re-score and sort the scorable prompts descending, then append the
        unscored tail
    \ENDIF
    \STATE Remove the first $\min(B,|\mathcal{Q}|)$ prompts as batch
    $\mathcal{B}$
    \STATE Generate independent on-policy training rollouts for $\mathcal{B}$
    and update the student with $\mathcal{L}_{\mathrm{OPD}}$
    \STATE $k\leftarrow k+1$
\ENDWHILE
\end{algorithmic}
\end{algorithm}

Static ReOrder-OPD adds exactly $|\mathcal{D}|$ student scoring rollouts.
If refreshes occur after fractions
$0<\alpha_1<\cdots<\alpha_J<1$ of the queue have been consumed, refresh $j$
scores roughly $(1-\alpha_j)|\mathcal{D}|$ remaining prompts, for about
$|\mathcal{D}|[1+\sum_j(1-\alpha_j)]$ scoring rollouts in total. Each prompt
is compared with at most $K_t$ correct teacher responses. Both variants reuse
the fixed teacher library; refreshes add student inference but no teacher
generation or OPD updates.

\section{Experiments}

\subsection{Experimental Setup}

\paragraph{Models.}
Our main teacher is Qwen3-30B-A3B-Instruct-2507, paired with Qwen3-1.7B, 4B,
and 8B students \citep{yang2025qwen3}. We also pair Gemma4-26B-A4B-it with
Gemma4-E2B and E4B students
\citep{team2026gemma}.

\paragraph{Datasets and teacher libraries.}
We use DeepMath-1K for reliability and ordering studies, DeepMath-5K for the
main comparison, and DeepMath-17K for data scaling. All sets are derived from
DeepMath-103K \citep{he2025deepmath}. For each prompt, the teacher library
contains up to 16 verifier-correct responses. Every student--seed pair
independently generates one rollout per prompt for proxy scoring. On
DeepMath-5K, 4,720 of 5,000 prompts have a nonempty library; the remaining 280
form the unscored tail and remain in the training pool.
Code-domain validation uses a 5,000-prompt subset of the code split of
Eurus-2-RL-Data \citep{cui2025process}.

\paragraph{Reliability diagnostic and scoring cost.}
On DeepMath-1K, $\widehat{R}$ uses $K_s=16$ student trajectories, five uniformly
spaced nonterminal prefixes per trajectory (at $1/6,\ldots,5/6$ of the
response), and $M=16$ teacher continuations per prefix: 80 tested prefixes and
1,280 continuations per prompt. Initial ReOrder scoring instead draws 16
teacher responses per prompt, verifies them to retain correct references, and
adds one independently sampled student scoring rollout per prompt. This rollout
is excluded from the $K_s=16$ trajectories used to estimate $\widehat{R}$ and
is never reused for OPD training. These preprocessing passes are absent from
vanilla OPD; matched comparisons equalize prompt visits and update counts, not
total generation and verification cost. The teacher library is
reusable across students, seeds, OPD objectives, and refreshes for a fixed
teacher--dataset pair; changing either requires rebuilding it.
Relative to the diagnostic $\widehat{R}$ estimate, initial scoring uses
$80\times$ fewer teacher samples and $16\times$ fewer student samples in
generated-sequence counts.

\paragraph{Benchmarks and metrics.}
We evaluate AIME24/25/26 and HMMT25-Feb, HMMT25-Nov, and HMMT26-Feb. For each
problem, mean@16 is verifier correctness averaged over 16 sampled responses.
The six-set score averages problem-level mean@16 over all 183 problems.
Standard deviations are computed across three training seeds. We use
mean@16 as the primary metric because it estimates expected correctness under
the fixed sampling configuration using all generated responses.
For code generation, we evaluate HumanEval+ (164 tasks) and MBPP+ (378 tasks)
with EvalPlus \citep{liu2023your}, and LiveCodeBench V6 (175 tasks)
\citep{jain2024livecodebench}. Code Avg.\ is their unweighted mean.

\paragraph{Hyperparameters and evaluation protocol.}
The 1K, 5K, and 17K comparisons use batch sizes 16, 64, and 64 for 63, 78,
and 266 updates, respectively, giving approximately one pass over each prompt
pool; Code-5K likewise uses batch size 64 for 78 updates. Vanilla OPD and
ReOrder use a learning rate of $10^{-6}$, one on-policy
student rollout per prompt visit, and a maximum response length of 16,384
tokens. Evaluation samples 16 responses with temperature 1.0 and top-$p$ 1.0.
Matched conditions share the prompt pool, update budget, and all other
settings; only prompt order differs. Three-seed studies use seeds 42--44.

\subsection{Main Results}

Across the main comparisons, ReOrder improves the matched aggregate for every
tested student and transfers from mathematics to code across all three training
seeds. A separate composition study shows gains on top of FiRe-OPD and ExOPD
at every tested student scale.

\paragraph{ReOrder-OPD improves mathematical reasoning across models and
benchmarks.}
Table~\ref{tab:main-results} compares ReOrder-OPD with matched vanilla OPD
across Qwen3 and Gemma4 teachers, five student configurations, six mathematics
benchmarks, and three training seeds. Within each student, the teacher,
initialization, prompt pool, OPD objective, update budget, and evaluation
protocol are held fixed, so the comparison isolates prompt order. ReOrder
raises the Six-set aggregate for all five students, with gains of
1.09--2.58 percentage points. All 15 model--seed aggregate paired differences
are positive, and 27 of the 30 model--benchmark means improve. The limited
benchmark-level regressions are localized rather than systematic across a
model family or benchmark group. The consistent aggregate gains across two
model families and five student scales show that reliability-aware ordering
improves performance under a fixed OPD update budget.

\begin{table*}[t!]
\centering
\small
\renewcommand{\arraystretch}{1.08}
\setlength{\tabcolsep}{4.0pt}
\begin{tabular}{@{}llcccccccc@{}}
\toprule
Student & Method & \multicolumn{3}{c}{AIME} &
\multicolumn{3}{c}{HMMT} & Six-set & $\Delta$ \\
\cmidrule(lr){3-5}\cmidrule(lr){6-8}
& & 2024 & 2025 & 2026 & 25-Feb & 25-Nov & 26-Feb & & \\
\midrule
\multicolumn{10}{l}{\textit{Qwen3 family}} \\
\textbf{1.7B} & Vanilla OPD &
\meanstd{33.47}{0.32} & \meanstd{27.36}{0.52} &
\meanstd{24.79}{1.67} & \bestmeanstd{16.67}{0.36} &
\meanstd{15.97}{0.43} & \meanstd{19.07}{0.87} &
\meanstd{22.83}{0.23} & -- \\
& ReOrder-OPD &
\bestmeanstd{34.31}{1.15} & \bestmeanstd{29.86}{1.34} &
\bestmeanstd{26.60}{1.05} & \meanstd{15.49}{1.05} &
\bestmeanstd{17.43}{1.77} & \bestmeanstd{20.20}{1.90} &
\bestmeanstd{23.92}{0.81} & $\mathbf{+1.09}$ \\
\addlinespace[2pt]
\textbf{4B} & Vanilla OPD &
\meanstd{53.89}{0.43} & \meanstd{47.85}{2.49} &
\meanstd{49.51}{0.84} & \meanstd{28.82}{0.64} &
\meanstd{33.89}{0.98} & \meanstd{29.48}{0.61} &
\meanstd{40.39}{0.31} & -- \\
& ReOrder-OPD &
\bestmeanstd{57.01}{2.30} & \bestmeanstd{48.89}{0.96} &
\bestmeanstd{50.42}{1.25} & \bestmeanstd{30.42}{1.10} &
\bestmeanstd{35.83}{1.27} & \bestmeanstd{30.74}{0.29} &
\bestmeanstd{42.03}{0.26} & $\mathbf{+1.64}$ \\
\addlinespace[2pt]
\textbf{8B} & Vanilla OPD &
\bestmeanstd{59.79}{2.35} & \meanstd{49.31}{1.94} &
\meanstd{52.71}{2.61} & \bestmeanstd{28.68}{0.43} &
\meanstd{39.10}{0.64} & \meanstd{31.57}{0.55} &
\meanstd{43.33}{1.02} & -- \\
& ReOrder-OPD &
\meanstd{59.51}{1.58} & \bestmeanstd{51.04}{1.25} &
\bestmeanstd{56.53}{0.48} & \meanstd{27.57}{1.68} &
\bestmeanstd{39.38}{0.55} & \bestmeanstd{33.78}{0.61} &
\bestmeanstd{44.46}{0.41} & $\mathbf{+1.13}$ \\
\midrule
\multicolumn{10}{l}{\textit{Gemma4 family}} \\
\textbf{E2B} & Vanilla OPD &
\meanstd{29.51}{2.09} & \meanstd{24.58}{0.36} &
\meanstd{25.49}{3.47} & \meanstd{11.25}{0.72} &
\meanstd{9.65}{1.18} & \meanstd{15.03}{1.79} &
\meanstd{19.18}{1.31} & -- \\
& ReOrder-OPD &
\bestmeanstd{33.75}{1.30} & \bestmeanstd{27.85}{0.67} &
\bestmeanstd{28.47}{2.44} & \bestmeanstd{11.88}{0.75} &
\bestmeanstd{12.50}{1.37} & \bestmeanstd{16.67}{0.87} &
\bestmeanstd{21.77}{0.96} & $\mathbf{+2.58}$ \\
\addlinespace[2pt]
\textbf{E4B} & Vanilla OPD &
\meanstd{45.42}{0.84} & \meanstd{31.74}{0.84} &
\meanstd{36.39}{1.05} & \meanstd{16.04}{0.36} &
\meanstd{22.99}{0.52} & \meanstd{25.57}{2.37} &
\meanstd{29.62}{0.68} & -- \\
& ReOrder-OPD &
\bestmeanstd{46.46}{2.66} & \bestmeanstd{34.38}{1.27} &
\bestmeanstd{39.10}{2.42} & \bestmeanstd{16.95}{1.07} &
\bestmeanstd{25.21}{2.05} & \bestmeanstd{26.14}{1.14} &
\bestmeanstd{31.28}{0.65} & $\mathbf{+1.66}$ \\
\bottomrule
\end{tabular}
\caption{Mathematical reasoning results of Vanilla OPD and ReOrder-OPD on
DeepMath-5K. We report mean@16 averaged over three training seeds, with sample
standard deviations in smaller type. Bold marks the better result within each
matched student configuration. $\Delta$ is the Six-set ReOrder-minus-Vanilla
difference computed from the unrounded aggregate.}
\label{tab:main-results}
\end{table*}

\paragraph{ReOrder-OPD generalizes to code generation.}
Table~\ref{tab:code-domain} tests the same ordering intervention on a different
class of verifiable generation tasks under matched models, prompt pools,
objectives, and update budgets. Code Avg.\ improves for both Qwen3-1.7B and
Qwen3-4B, and all six student--seed paired differences are positive. The mean
direction is also positive on HumanEval+, MBPP+, and LiveCodeBench V6 for both
students. Prompt ordering benefits code generation without changing
the underlying OPD objective, extending the scheduling principle beyond
mathematical reasoning.

\begin{table}[t]
\centering
\small
\renewcommand{\arraystretch}{1.08}
\setlength{\tabcolsep}{1.0pt}
\begin{tabular*}{\columnwidth}{@{\extracolsep{\fill}}ll@{\hspace{3pt}}*{4}{c}@{}}
\toprule
Student & Method & HE+ & MBPP+ & LCB-v6 & Avg. \\
\midrule
\textbf{1.7B} & Vanilla &
\meanstd{68.20}{0.43} & \meanstd{59.15}{0.21} &
\meanstd{17.99}{0.39} & \meanstd{48.45}{0.23} \\
& ReOrder\hspace{3pt} &
\bestmeanstd{69.05}{0.40} & \bestmeanstd{59.54}{0.54} &
\bestmeanstd{19.23}{0.42} & \bestmeanstd{49.27}{0.13} \\
\addlinespace[1.2pt]
\textbf{4B} & Vanilla &
\meanstd{81.80}{0.87} & \meanstd{71.25}{0.32} &
\meanstd{26.83}{0.52} & \meanstd{59.96}{0.41} \\
& ReOrder\hspace{3pt} &
\bestmeanstd{83.38}{0.04} & \bestmeanstd{72.20}{0.48} &
\bestmeanstd{28.32}{0.41} & \bestmeanstd{61.30}{0.30} \\
\bottomrule
\end{tabular*}
\caption{Code-generation mean@16 (\%) for Qwen3 students. Values are mean
$\pm$ sample SD over seeds 42--44. HE+ and LCB-v6 denote HumanEval+ and
LiveCodeBench V6; Avg.\ equally weights the three benchmarks. Bold marks the
better result within each matched student configuration.}
\label{tab:code-domain}
\end{table}

\paragraph{ReOrder complements trajectory-level OPD methods.}
FiRe-OPD and ExOPD \citep{li2026filter,yang2026learning} determine how
supervision is applied within an on-policy rollout, whereas ReOrder determines
when the prompt is visited before that rollout is generated.
Table~\ref{tab:composition} composes the two intervention levels while retaining
each trajectory-level objective, teacher, prompt pool, and update budget. At
seed 42, adding ReOrder improves all six combinations spanning three student
scales and both objectives. The gains appear for both FiRe-OPD and ExOPD at
every tested scale, rather than depending on one particular trajectory-level
loss. Improvements in all six tested combinations demonstrate that prompt
scheduling is compatible with both trajectory-level objectives across the
three student scales.

\begin{table}[t]
\centering
\small
\renewcommand{\arraystretch}{1.04}
\setlength{\tabcolsep}{3pt}
\begin{tabular}{@{}llrrr@{}}
\toprule
Student & Objective & Base & +ReOrder & $\Delta$ \\
\midrule
Qwen3-1.7B & FiRe & 24.38 & \textbf{27.71} & +3.33 \\
            & ExOPD & 26.88 & \textbf{28.23} & +1.35 \\
Qwen3-4B   & FiRe & 48.02 & \textbf{49.79} & +1.77 \\
            & ExOPD & 52.40 & \textbf{53.44} & +1.04 \\
Qwen3-8B   & FiRe & 50.73 & \textbf{52.92} & +2.19 \\
            & ExOPD & 55.31 & \textbf{56.56} & +1.25 \\
\bottomrule
\end{tabular}
\caption{Prompt-level ReOrder composed with trajectory-level FiRe-OPD and
ExOPD on DeepMath-5K under seed 42 (AIME25/26 mean@16, \%).}
\label{tab:composition}
\end{table}

\subsection{Reliability and Proxy Analysis}

\paragraph{Teacher continuation reliability separates prompt groups by OPD utility.}
The first block of Table~\ref{tab:oracle-r} compares equal-size subsets drawn
from the same initial student state, with the subset size and 21-update training
budget held fixed. High-$R$ prompts obtain the strongest average result and
exceed low-$R$ prompts by 1.25 points on the combined score. Middle-$R$ and
random are close, so the useful distinction is primarily between coarse
reliability strata rather than a finely resolved ranking of individual prompt
gains. These results show that $R$ separates high- and low-utility prompt
groups at the coarse resolution needed for scheduling.

\paragraph{Reliability-aware ordering outperforms alternative curricula.}
The full-pool block of Table~\ref{tab:oracle-r} changes only access order:
descending $R$ is 1.94 points above vanilla, ascending $R$ is 1.94 below it,
and all three seed-level differences favor descending $R$.
Figure~\ref{fig:motivation}(b) holds the prompt pool, objective, and update
budget fixed. Student-correct-first, teacher-success-first, and annotated
easy-first sort prompts respectively by the base student's empirical
correctness rate, the teacher's verified success rate, and the dataset
difficulty annotation; ReOrder outperforms all three orders. $R$ summarizes the
interaction among the prompt, student-induced prefix, and teacher continuation,
and is the strongest scheduling signal among the tested alternatives.

\begin{table}[t!]
\centering
\small
\renewcommand{\arraystretch}{1.08}
\setlength{\tabcolsep}{3.0pt}
\begin{tabular}{@{}lccc@{}}
\toprule
\multicolumn{4}{c}{Equal-size oracle-$R$ subsets} \\
Subset & AIME25 & AIME26 & Combined \\
\midrule
Low-$R$ & \meanstd{15.42}{0.32} & \meanstd{12.71}{1.16} & \meanstd{14.06}{0.68} \\
Middle-$R$ & \meanstd{15.83}{0.87} & \meanstd{13.54}{1.18} & \meanstd{14.69}{1.03} \\
High-$R$ & \bestmeanstd{16.46}{0.83} & \bestmeanstd{14.17}{0.95} & \bestmeanstd{15.31}{0.81} \\
Random & \meanstd{15.63}{0.87} & \meanstd{13.75}{0.64} & \meanstd{14.69}{0.72} \\
\midrule
\multicolumn{4}{c}{Full-pool oracle-$R$ ordering} \\
Order & \multicolumn{2}{c}{AIME25/26} & $\Delta$ vs.\ vanilla \\
\midrule
$R$ ascending & \multicolumn{2}{c}{\meanstd{23.13}{1.25}} & $-1.94$ \\
Vanilla OPD & \multicolumn{2}{c}{\meanstd{25.07}{0.84}} & -- \\
$R$ descending & \multicolumn{2}{c}{\bestmeanstd{27.01}{1.08}} &
$\mathbf{+1.94}$ \\
\bottomrule
\end{tabular}
\caption{DeepMath-1K oracle-$R$ experiments. The first block trains for 21
updates on 336-prompt low, middle, high, or random subsets; subsets may overlap.
The second block retains the full pool and changes its order. Values are
mean@16 (\%), reported as mean $\pm$ sample SD over three seeds.}
\label{tab:oracle-r}
\end{table}

\begin{figure}[t]
\centering
\includegraphics[width=\columnwidth]{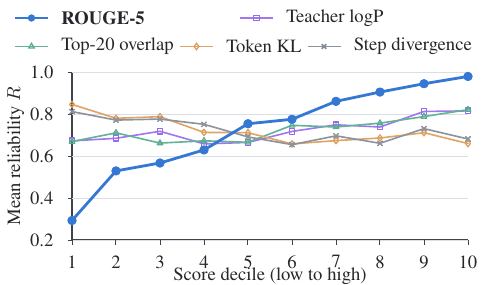}
\caption{Mean diagnostic $R$ across ten equal-frequency bins formed
independently by each score on the same 905 prompts. Bins increase with each
score; token KL and step divergence are expected to decrease with reliability.}
\label{fig:proxy-deciles}
\end{figure}

\paragraph{One-rollout ROUGE-5 separates coarse reliability strata.}
We evaluate whether the inexpensive score used by ReOrder preserves the coarse
reliability structure exposed by the oracle experiments. On the same 905
prompts, the actual one-rollout ROUGE-5 score in
Figure~\ref{fig:proxy-deciles} partitions prompts into deciles whose mean $R$
rises from 0.29 to 0.98, with all nine adjacent changes positive. The local
signals in Figure~\ref{fig:motivation}(a) have weak prompt-level associations
with continuation success, and their decile curves reverse direction multiple
times. ROUGE-5 thus serves as an ordinal scheduling score that separates
coarse reliability strata, rather than a calibrated estimator of $R$.
Maximum aggregation accounts for \mbox{multiple} valid solution paths by requiring
the student rollout to resemble at least one verifier-correct teacher
trajectory.

\paragraph{Ablation of the proxy design.}
Table~\ref{tab:proxy-ablation} compares proxy variants on the same 905 prompts,
fixed scoring rollouts, and teacher trajectories. ROUGE-3 and ROUGE-5 both
produce nine increasing decile transitions, whereas ROUGE-1 produces eight.
Their agreement indicates that the coarse ranking is not specific to one
$n$-gram order.
With ROUGE-5 fixed, maximum aggregation preserves all nine increases and
produces a larger high--low reliability separation than either mean or
random-teacher aggregation on the shared rollouts.
Maximum assigns high priority when the student resembles any one correct
solution, which is appropriate for prompts with multiple valid paths.

\begin{table}[t]
\centering
\small
\renewcommand{\arraystretch}{1.02}
\setlength{\tabcolsep}{3.2pt}
\begin{tabular}{@{}lccc@{}}
\toprule
Design & Decile $\rho$ & Up & $\Delta R$ \\
\midrule
ROUGE-1 / max & 0.964 & 8/9 & 0.624 \\
ROUGE-3 / max & 1.000 & 9/9 & 0.703 \\
ROUGE-5 / max & 1.000 & 9/9 & 0.686 \\
ROUGE-5 / mean & 0.988 & 8/9 & 0.558 \\
ROUGE-5 / random & $0.992{\pm}0.007$ & 8.3/9 &
$0.508{\pm}0.022$ \\
\bottomrule
\end{tabular}
\caption{Proxy-design ablations on the common 905 prompts. Decile $\rho$ is
the Spearman correlation between decile index and mean $R$; Up counts
increasing adjacent deciles, and $\Delta R$ is the highest--lowest mean
reliability. Random-teacher values are mean $\pm$ sample SD over 20 repetitions.}
\label{tab:proxy-ablation}
\end{table}

Oracle $R$ identifies the useful access direction, and one scoring rollout
recovers it at the coarse resolution needed for scheduling. Training still
draws fresh on-policy trajectories and applies the original OPD objective.

\subsection{Further Analysis}

\paragraph{ReOrder remains effective at larger data scales.}
Table~\ref{tab:data-scale} tests whether a static ordering remains useful as
the DeepMath prompt pool grows. Within each scale, Vanilla and ReOrder use the
same data, student, objective, and matched number of updates. ReOrder improves
the AIME25/26 score at 1K, 5K, and 17K, showing that the benefit persists
beyond the small-pool setting. The gain decreases from 3.33 to 1.15 points as
the pool and training trajectory grow. The shrinking gain suggests that a
static ranking becomes less informative over a longer training trajectory,
motivating priority refresh.

\begin{table}[t]
\centering
\small
\renewcommand{\arraystretch}{1.05}
\setlength{\tabcolsep}{4.0pt}
\begin{tabular}{@{}lrrrr@{}}
\toprule
Data & Updates & Vanilla & ReOrder & $\Delta$ \\
\midrule
1K & 63  & 24.17 & \textbf{27.50} & +3.33 \\
5K & 78  & 25.83 & \textbf{28.44} & +2.60 \\
17K & 266 & 26.88 & \textbf{28.02} & +1.15 \\
\bottomrule
\end{tabular}
\caption{Static ordering across DeepMath data scales for Qwen3-1.7B under
seed 42. Values are AIME25/26 mean@16 (\%); each comparison uses the matched
update count shown.}
\label{tab:data-scale}
\end{table}

\paragraph{Dynamic refresh improves static ordering.}
The scale trend motivates refreshing priorities as the student evolves.
Table~\ref{tab:dynamic} compares Vanilla, Static ReOrder, and Dynamic ReOrder
on DeepMath-17K under the same teacher library, prompt pool, and 266-update
budget. Dynamic refresh changes only scores and order for unvisited prompts.
For every student, Dynamic has the highest combined score, followed by Static
and then Vanilla. The individual columns contain
two exceptions: Static is highest on AIME26 for Qwen3-4B, and Vanilla is
highest there for Qwen3-8B. The aggregate pattern matches the policy
dependence of $q(x;\pi_s)$: refreshed priorities track the student generating
the remaining on-policy trajectories. Dynamic ReOrder adds student-side
scoring but no teacher generation or OPD updates at refresh time, preserving
the matched OPD update budget.

\begin{table}[t]
\centering
\small
\renewcommand{\arraystretch}{1.05}
\setlength{\tabcolsep}{3.0pt}
\begin{tabular}{@{}llrrr@{}}
\toprule
Student & Method & AIME25 & AIME26 & Combined \\
\midrule
Qwen3-1.7B & Vanilla & 27.71 & 26.04 & 26.88 \\
            & Static  & 27.50 & 28.54 & 28.02 \\
            & Dynamic & \textbf{28.13} & \textbf{29.38} & \textbf{28.75} \\
\addlinespace[1.2pt]
Qwen3-4B   & Vanilla & 47.29 & 49.79 & 48.54 \\
            & Static  & 47.29 & \textbf{53.13} & 50.21 \\
            & Dynamic & \textbf{49.17} & 52.71 & \textbf{50.94} \\
\addlinespace[1.2pt]
Qwen3-8B   & Vanilla & 47.50 & \textbf{56.88} & 52.19 \\
            & Static  & 48.96 & 56.67 & 52.81 \\
            & Dynamic & \textbf{51.88} & 56.04 & \textbf{53.96} \\
\bottomrule
\end{tabular}
\caption{Static and dynamic ordering on DeepMath-17K under seed 42. Values
are mean@16 (\%) after 266 updates. Dynamic ReOrder refreshes unvisited-prompt
scores after updates 133, 177, and 222. Bold marks the best result within each
student and benchmark.}
\label{tab:dynamic}
\end{table}

\section{Conclusion}

ReOrder-OPD schedules prompts using a one-rollout proxy for teacher
continuation reliability $R$, which measures whether the teacher can correctly
complete a student-induced prefix.
Oracle experiments show that $R$ separates prompts with different OPD utility
and favor high-$R$-first access. One scoring rollout supplies the coarse
ordering; fresh on-policy rollouts supply the training trajectories.

ReOrder improves matched aggregates across two model families, data scales,
mathematics, and code, and adds gains to FiRe-OPD and ExOPD in all six tested
combinations. Prompt scheduling and trajectory-level supervision intervene
differently: ReOrder chooses when to visit a prompt, whereas the OPD objective
controls how its rollout is trained. Ordering matters because each update
changes the policy that generates later trajectories. Static gains show the
initial ranking is useful; dynamic refresh shows priority evolves with the
student rather than the dataset alone. Verifier-correct libraries remain
necessary, and refresh adds scoring cost.

\bibliography{aaai2027_v5}

@article{hinton2015distilling,
  title={Distilling the knowledge in a neural network},
  author={Hinton, Geoffrey and Vinyals, Oriol and Dean, Jeff},
  journal={arXiv preprint arXiv:1503.02531},
  year={2015}
}

@inproceedings{ross2011reduction,
  title={A reduction of imitation learning and structured prediction to no-regret online learning},
  author={Ross, St{\'e}phane and Gordon, Geoffrey and Bagnell, Drew},
  booktitle={Proceedings of the fourteenth international conference on artificial intelligence and statistics},
  pages={627--635},
  year={2011},
  organization={JMLR Workshop and Conference Proceedings}
}

@article{bengio2015scheduled,
  title={Scheduled sampling for sequence prediction with recurrent neural networks},
  author={Bengio, Samy and Vinyals, Oriol and Jaitly, Navdeep and Shazeer, Noam},
  journal={Advances in neural information processing systems},
  volume={28},
  year={2015}
}

@inproceedings{kim2016sequence,
  title={Sequence-level knowledge distillation},
  author={Kim, Yoon and Rush, Alexander M},
  booktitle={Proceedings of the 2016 conference on empirical methods in natural language processing},
  pages={1317--1327},
  year={2016}
}

@inproceedings{gu2024minillm,
  title={Minillm: Knowledge distillation of large language models},
  author={Gu, Yuxian and Dong, Li and Wei, Furu and Huang, Minlie},
  booktitle={The twelfth international conference on learning representations},
  year={2024}
}

@inproceedings{agarwal2024policy,
  title={On-policy distillation of language models: Learning from self-generated mistakes},
  author={Agarwal, Rishabh and Vieillard, Nino and Zhou, Yongchao and Stanczyk, Piotr and Ramos Garea, Sabela and Geist, Matthieu and Bachem, Olivier},
  booktitle={International Conference on Learning Representations},
  volume={2024},
  pages={21246--21263},
  year={2024}
}

@article{lee2026sead,
  title={SEAD: Competence-Aware On-Policy Distillation via Entropy-Guided Supervision},
  author={Lee, Chia-Hsuan and Cheng, Zelei and Wang, Yu and Ni, Renkun and Sahu, Sambit and Zhang, Shi-Xiong and Campbell, William},
  journal={arXiv preprint arXiv:2606.28562},
  year={2026}
}

@article{team2026gemma,
  title={Gemma 4 technical report},
  author={Team, Gemma and Abd, Sherif El and Aggarwal, Vaibhav and Algayres, Robin and Andreev, Alek and Bachem, Olivier and Ballantyne, Ian and Brick, Cormac and C{\u{a}}rbune, Victor and Casbon, Michelle and others},
  journal={arXiv preprint arXiv:2607.02770},
  year={2026}
}

@article{li2026rethinking,
  title={Rethinking on-policy distillation of large language models: Phenomenology, mechanism, and recipe},
  author={Li, Yaxuan and Zuo, Yuxin and He, Bingxiang and Zhang, Jinqian and Xiao, Chaojun and Qian, Cheng and Yu, Tianyu and Gao, Huan-ang and Yang, Wenkai and Liu, Zhiyuan and others},
  journal={arXiv preprint arXiv:2604.13016},
  year={2026}
}

@article{fu2026revisiting,
  title={Revisiting on-policy distillation: Empirical failure modes and simple fixes},
  author={Fu, Yuqian and Huang, Haohuan and Jiang, Kaiwen and Liu, Jiacai and Jiang, Zhuo and Zhu, Yuanheng and Zhao, Dongbin},
  journal={arXiv preprint arXiv:2603.25562},
  year={2026}
}

@article{zhong2026sod,
  title={Sod: Step-wise on-policy distillation for small language model agents},
  author={Zhong, Qiyong and Zheng, Mao and Song, Mingyang and Lin, Xin and Sun, Jie and Jiang, Houcheng and Wang, Xiang and Fang, Junfeng},
  journal={arXiv preprint arXiv:2605.07725},
  year={2026}
}

@article{li2026filter,
  title={Filter, then reweight: Rethinking optimization granularity in on-policy distillation},
  author={Li, Yuying and Zheng, Leqi and Yu, Yongzi and Zhou, Wenrui and Zhong, Xuchang and Hu, Xing and Jin, Jing and Yuan, Hangjie and Feng, Tao},
  journal={arXiv preprint arXiv:2606.02684},
  year={2026}
}

@article{yang2026prune,
  title={Prune-OPD: Efficient and Reliable On-Policy Distillation for Long-Horizon Reasoning},
  author={Yang, Zhicheng and Guo, Zhijiang and Song, Yifan and Xu, Minrui and Wang, Yongxin and Wang, Yiwei and Liang, Xiaodan and Tang, Jing},
  journal={arXiv preprint arXiv:2605.07804},
  year={2026}
}

@article{li2026policy,
  title={On-Policy Distillation with Curriculum Turn-level Guidance for Multi-turn Agents},
  author={Li, Gengsheng and Zheng, Mao and Song, Mingyang and Liu, Ruiqi and Yang, Tianyu and Sun, Jie and Zhong, Qiyong and Guo, Haiyun and Fang, Junfeng and Zhang, Dan and others},
  journal={arXiv preprint arXiv:2606.15912},
  year={2026}
}

@article{yang2026learning,
  title={Learning beyond teacher: Generalized on-policy distillation with reward extrapolation},
  author={Yang, Wenkai and Liu, Weijie and Xie, Ruobing and Yang, Kai and Yang, Saiyong and Lin, Yankai},
  journal={arXiv preprint arXiv:2602.12125},
  year={2026}
}

@article{yang2025qwen3,
  title={Qwen3 technical report},
  author={Yang, An and Li, Anfeng and Yang, Baosong and Zhang, Beichen and Hui, Binyuan and Zheng, Bo and Yu, Bowen and Gao, Chang and Huang, Chengen and Lv, Chenxu and others},
  journal={arXiv preprint arXiv:2505.09388},
  year={2025}
}

@article{he2025deepmath,
  title={Deepmath-103k: A large-scale, challenging, decontaminated, and verifiable mathematical dataset for advancing reasoning},
  author={He, Zhiwei and Liang, Tian and Xu, Jiahao and Liu, Qiuzhi and Chen, Xingyu and Wang, Yue and Song, Linfeng and Yu, Dian and Liang, Zhenwen and Wang, Wenxuan and others},
  journal={arXiv preprint arXiv:2504.11456},
  year={2025}
}

@article{cui2025process,
  title={Process reinforcement through implicit rewards},
  author={Cui, Ganqu and Yuan, Lifan and Wang, Zefan and Wang, Hanbin and Zhang, Yuchen and Chen, Jiacheng and Li, Wendi and He, Bingxiang and Fan, Yuchen and Yu, Tianyu and others},
  journal={arXiv preprint arXiv:2502.01456},
  year={2025}
}

@article{liu2023your,
  title={Is your code generated by chatgpt really correct? rigorous evaluation of large language models for code generation},
  author={Liu, Jiawei and Xia, Chunqiu Steven and Wang, Yuyao and Zhang, Lingming},
  journal={Advances in neural information processing systems},
  volume={36},
  pages={21558--21572},
  year={2023}
}

@article{jain2024livecodebench,
  title={Livecodebench: Holistic and contamination free evaluation of large language models for code},
  author={Jain, Naman and Han, King and Gu, Alex and Li, Wen-Ding and Yan, Fanjia and Zhang, Tianjun and Wang, Sida and Solar-Lezama, Armando and Sen, Koushik and Stoica, Ion},
  journal={arXiv preprint arXiv:2403.07974},
  year={2024}
}

@inproceedings{lin2004rouge,
  title={Rouge: A package for automatic evaluation of summaries},
  author={Lin, Chin-Yew},
  booktitle={Text summarization branches out},
  pages={74--81},
  year={2004}
}

@inproceedings{kong2026rethinking,
  title={Rethinking the sampling criteria in reinforcement learning for LLM reasoning: A competence-difficulty alignment perspective},
  author={Kong, Deyang and Guo, Qi and Xi, Xiangyu and Wang, Wei and Wang, Jingang and Cai, Xunliang and Zhang, Shikun and Ye, Wei},
  booktitle={Proceedings of the AAAI Conference on Artificial Intelligence},
  volume={40},
  number={37},
  pages={31438--31446},
  year={2026}
}

@inproceedings{bengio2009curriculum,
  title={Curriculum learning},
  author={Bengio, Yoshua and Louradour, J{\'e}r{\^o}me and Collobert, Ronan and Weston, Jason},
  booktitle={Proceedings of the 26th annual international conference on machine learning},
  pages={41--48},
  year={2009}
}

@article{kumar2010self,
  title={Self-paced learning for latent variable models},
  author={Kumar, M and Packer, Benjamin and Koller, Daphne},
  journal={Advances in neural information processing systems},
  volume={23},
  year={2010}
}

@article{liu2026teacher,
  title={When Are Teacher Tokens Reliable? Position-Weighted On-Policy Self-Distillation for Reasoning},
  author={Liu, Xiaogeng and Wang, Xinyan and Ma, Yingzi and Zhang, Yechao and Xiao, Chaowei},
  journal={arXiv preprint arXiv:2605.21606},
  year={2026}
}

@article{zhao2026prefix,
  title={Prefix-Guided On-Policy Distillation: Mining Golden Trajectories from Rollouts},
  author={Zhao, Qingfei and Song, Huan and Tian, Shuyu and Shao, Jiawei and Li, Xuelong},
  journal={arXiv preprint arXiv:2606.21994},
  year={2026}
}

@article{akhondzadeh2026reward,
  title={Reward-Gated On-Policy Distillation},
  author={Akhondzadeh, Mohammad Sadegh and Lingam, Vijay and Tejaswi, Atula and Ekbote, Chanakya and Sanghavi, Sujay and Bojchevski, Aleksandar},
  journal={arXiv preprint arXiv:2607.04037},
  year={2026}
}

@inproceedings{jia2026makes,
  title={What makes a good curriculum? disentangling the effects of data ordering on llm mathematical reasoning},
  author={Jia, Yaning and Zhang, Chunhui and Diao, Xingjian and Yuan, Xiangchi and Ouyang, Zhongyu and Ma, Chiyu and Vosoughi, Soroush},
  booktitle={Proceedings of the 64th Annual Meeting of the Association for Computational Linguistics (Volume 1: Long Papers)},
  pages={34472--34488},
  year={2026}
}

@article{gou2021knowledge,
  title={Knowledge distillation: A survey},
  author={Gou, Jianping and Yu, Baosheng and Maybank, Stephen J and Tao, Dacheng},
  journal={International journal of computer vision},
  volume={129},
  number={6},
  pages={1789--1819},
  year={2021},
  publisher={Springer}
}

@inproceedings{graves2017automated,
  title={Automated curriculum learning for neural networks},
  author={Graves, Alex and Bellemare, Marc G and Menick, Jacob and Munos, Remi and Kavukcuoglu, Koray},
  booktitle={international conference on machine learning},
  pages={1311--1320},
  year={2017},
  organization={Pmlr}
}

@inproceedings{jiang2018mentornet,
  title={Mentornet: Learning data-driven curriculum for very deep neural networks on corrupted labels},
  author={Jiang, Lu and Zhou, Zhengyuan and Leung, Thomas and Li, Li-Jia and Fei-Fei, Li},
  booktitle={International conference on machine learning},
  pages={2304--2313},
  year={2018},
  organization={PMLR}
}

@inproceedings{hacohen2019power,
  title={On the power of curriculum learning in training deep networks},
  author={Hacohen, Guy and Weinshall, Daphna},
  booktitle={International conference on machine learning},
  pages={2535--2544},
  year={2019},
  organization={PMLR}
}

@inproceedings{platanios2019competence,
  title={Competence-based curriculum learning for neural machine translation},
  author={Platanios, Emmanouil Antonios and Stretcu, Otilia and Neubig, Graham and Poczos, Barnabas and Mitchell, Tom},
  booktitle={Proceedings of the 2019 conference of the North American chapter of the association for computational linguistics: human language technologies, volume 1 (long and short papers)},
  pages={1162--1172},
  year={2019}
}

@article{soviany2022curriculum,
  title={Curriculum learning: A survey},
  author={Soviany, Petru and Ionescu, Radu Tudor and Rota, Paolo and Sebe, Nicu},
  journal={International Journal of Computer Vision},
  volume={130},
  number={6},
  pages={1526--1565},
  year={2022},
  publisher={Springer}
}

\end{document}